\documentclass{article}

\usepackage[final,nonatbib]{neurips_2022}

\usepackage[utf8]{inputenc} %
\usepackage[T1]{fontenc}    %
\usepackage[hidelinks]{hyperref}       %
\usepackage{url}            %
\usepackage{booktabs}       %
\usepackage{amsfonts}       %
\usepackage{nicefrac}       %
\usepackage{microtype}      %
\usepackage{xcolor}         %
\usepackage[pdftex]{graphicx} 
\usepackage{float}
\usepackage[algo2e, ruled, noline, nofillcomment]{algorithm2e}
\usepackage{subfig}
\usepackage{amsmath,amsfonts,amssymb,mathtools}
\usepackage[para,norule]{footmisc}

\usepackage[square,numbers]{natbib}

\newcommand{\Real}{\ensuremath{\mathbb{R}}}
\newcommand{\Nat}{\ensuremath{\mathbb{N}}}

\newcommand{\Bin}{\ensuremath{\left\lbrace0,1\right\rbrace}}

\DeclareMathOperator*{\argmin}{arg\,min}

\newcommand{\Transpose}{\ensuremath{\top}} %

\providecommand{\abs}[1]{\left\lvert#1\right\rvert}

\newtheorem{definition}{Definition}%
\newtheorem{problem}{Problem}%

\DeclareMathOperator*{\bigoh}{\mathcal{O}}

\newcommand{\Rom}[1]{\uppercase\expandafter{\romannumeral #1\relax}}
\newcommand{\rom}[1]{\lowercase\expandafter{\romannumeral #1\relax}}

\newcommand{\target}{\ensuremath{\mathrm{A}}}
\newcommand{\lhs}{\ensuremath{\mathrm{U}}} 
\newcommand{\rhs}{\ensuremath{\mathrm{V}}} 
\newcommand{\recons}{\ensuremath{\mathrm{B}}}

\newcommand{\prox}{\ensuremath{\operatorname{prox}}}
\newcommand{\sign}{\ensuremath{\operatorname{sign}}}

\newcommand{\pimp}{\textsc{Primp}\xspace}
\newcommand{\asso}{\textsc{Asso}\xspace}
\newcommand{\sofa}{\textsc{Sofa}\xspace}
\newcommand{\grecond}{\textsc{Grecond}\xspace}

\newcommand{\nmf}{\textsc{nmf}\xspace}
\newcommand{\binaps}{\textsc{Binaps}\xspace}
\newcommand{\ourmethod}{\textsc{Elbmf}\xspace}
\newcommand{\LOM}{\textsc{OrM}\xspace}

\newcommand{\ourregularizer}{{\scshape Elb}\xspace}
\usepackage{siunitx}
\usepackage{numprint}
\usepackage{comment}

\title{Efficiently Factorizing Boolean Matrices\\using Proximal Gradient Descent} 

\author{%
  Sebastian Dalleiger\\
  {\small CISPA Helmholtz Center for Information Security}\\
  {\small \href{sebastian.dalleiger@cispa.de}{sebastian.dalleiger@cispa.de}}
  \And
  Jilles Vreeken\\
  {\small CISPA Helmholtz Center for Information Security}\\
  {\small \href{jv@cispa.de}{jv@cispa.de}}
}

\newcommand{\oururl}{\href{https://doi.org/10.5281/zenodo.7187021}{doi.org/10.5281/zenodo.7187021}}

\newcommand{\ourdoiurl}{\href{https://doi.org/10.5281/zenodo.7187021}{10.5281/zenodo.7187021}}

\begin{document}

\maketitle

\begin{abstract}
Addressing the interpretability problem of NMF on Boolean data, Boolean Matrix Factorization (BMF) uses Boolean algebra to decompose the input into low-rank Boolean factor matrices.
These matrices are highly interpretable and very useful in practice,
but they come at the high computational cost of solving an NP-hard combinatorial optimization problem.
To reduce the computational burden, we propose to relax BMF continuously using a novel elastic-binary regularizer, from which we derive a proximal gradient algorithm.
Through an extensive set of experiments, we demonstrate that our method works well in practice:
On synthetic data, we show that it converges quickly, recovers the ground truth precisely, and estimates the simulated rank exactly.
On real-world data, we improve upon the state of the art in recall, loss, and runtime,
and a case study from the medical domain confirms that our results are easily interpretable and semantically meaningful. 

\end{abstract}

\section{Introduction}

Discovering groups in data and expressing them in terms of common concepts is a central problem in many scientific domains and business applications, including cancer genomics \cite{Liang:2020:BEM}, neuroscience \cite{Haddad:2018:Identifying}, and recommender systems \cite{Ignatov:2014:Boolean}.
This problem is often addressed using variants of \emph{matrix factorization}, a family of methods that decompose the target matrix into a set of typically low-rank factor matrices whose product approximates the input well. 
Prominent examples of matrix factorization are Singular Value Decomposition (SVD)~\cite{Golub:1996:Matrix}, Principal Component Analysis (PCA) \cite{Golub:1996:Matrix}, and Nonnegative Matrix Factorization (NMF) \cite{Paatero:1994:Positive,Lee:1999:Learning,Lee:2000:Algorithms}. 
These methods differ in how they constrain the matrices involved: 
SVD and PCA require orthogonal factors, while NMF constrains the target matrix and the factors to be nonnegative. 

SVD, PCA, and NMF achieve interpretable results---unless the data is Boolean, which is ubiquitous in the real world.
In this case, their results are hard to interpret directly because the input domain differs from the output domain, such that post-processing is required to extract useful information.
\emph{Boolean Matrix Factorization} (BMF) addresses this problem by seeking two low-rank \emph{Boolean} factor matrices whose Boolean product is close to the Boolean target matrix \cite{Miettinen:2008:discrete}. 
The output matrices, now lying in the same domain as the input, are interpretable and useful,
but they come at the computational cost of solving an NP-hard combinatorial optimization problem \cite{Orlin:1977:Contentment,Miettinen:2008:discrete,Miettinen:2020:Recent}. 
To make BMF applicable in practice, we need efficient approximation algorithms.

There are many ways to approximate BMF---%
for example, by exploiting its underlying combinatorial or spatial structure~\cite{Miettinen:2008:discrete,Belohlavek:2010:Discovery,Belohlavek:2018:New},
using probabilistic inference~\cite{Rukat:2017:interpretable,Rukat:2017:Bayesian,Rukat:2018:Probabilistic}, 
or solving the related Bi-Clustering problem~\cite{Neumann:2018:Bipartite, Neumann:2020:Biclustering}.
Although these approaches achieve impressive results, they fall short when the input data is large and noisy. 
Hence, we take a different approach to overcome BMF's computational barrier.
Starting from an NMF-like optimization problem, 
we derive a continuous relaxation of the original BMF formulation that allows intermediate solutions to be real-valued.
Inspired by the elastic-net regularizer~\cite{Zou:2005:Regularization}, we introduce the novel \emph{elastic binary (\ourregularizer) regularizer} to regularize towards Boolean factor matrices.
We obtain an efficient-to-compute \emph{proximal operator} from our \ourregularizer regularizer that projects relaxed real-valued factors towards being Boolean, 
which allows us to leverage fast gradient-based optimization procedures.
In stark contrast to the state of the art~\cite{Hess:2017:PRIMPING,Hess:2018:Trustworthy,Hess:2021:BROCCOLI},
which requires heavy post-hoc post-processing to actually achieve Boolean factors, 
we ensure a Boolean outcome upon convergence
by gradually increasing the projection strength using a \emph{regularization rate}.
We combine our relaxation, efficient proximal operator, and regularization rate into an \emph{Elastic Boolean Matrix Factorization} algorithm (\ourmethod) that 
scales to large data, results in accurate reconstructions, and does so without relying on heavy post-processing procedures.  
\ourregularizer and its rate are, however, not confined to BMF and can regularize, e.g., binary MF or bi-clustering~\cite{Hess:2021:BROCCOLI}.

In summary, our main contributions are as follows:
\begin{enumerate}
    \item We introduce the \ourregularizer regularizer.
    \item We overcome the computational hardness of BMF leveraging a novel relaxed BMF problem.
    \item We efficiently solve the relaxed BMF problem using an optimization algorithm based on proximal gradient descent.
\end{enumerate}

The remainder of the paper proceeds as follows.
In Sec.~\ref{sec:theory}, we formally introduce the BMF problem and its relaxation, 
define our \ourregularizer regularizer and its proximal point operator, and show how to ensure a Boolean outcome upon convergence. 
We discuss related work in Sec.~\ref{sec:related},
validate our method through an extensive set of experiments in Sec.~\ref{sec:experiments}, 
and conclude with a discussion in Sec.~\ref{sec:conclusion}.

\section{Theory}\label{sec:theory}

Our goal is to factorize a given Boolean target matrix into at least two smaller, low-rank Boolean factor 
matrices, 
whose product comes close to the target matrix. 
Since the factor matrices are Boolean, this product follows the algebra of a Boolean semi-ring, i.e., it is identical to the standard outer product on a field where addition obeys $1 + 1 = 1$.
We define the product between two Boolean matrices $\lhs \in \Bin^{n \times k}$ and $\rhs \in \Bin^{k \times m}$ on a Boolean semi-ring $(\Bin, \lor, \land)$ as
\begin{equation}
    [\lhs \circ \rhs]_{ij} = \bigvee_{l \in [k]} \lhs_{il} \rhs_{lj}\;,
    \label{eq:semiproduct}
\end{equation}
where $\lhs \in \Bin^{n \times k}$, $\rhs \in \Bin^{k \times m}$, and $\lhs \circ \rhs \in \Bin^{n \times m}$. 
This gives rise to the BMF problem.
\begin{problem}[Boolean Matrix Factorization]
    For a given target matrix $\target \in \Bin^{n \times m}$, a given matrix rank $\Nat \ni k \leq \min\{n, m\}$, and $A \oplus B$ denoting logical exclusive or, discover the factor matrices $\lhs \in \Bin^{n \times k}$ and $\rhs \in \Bin^{k \times m}$ that minimize
    \begin{equation}
        \| \target - \lhs \circ \rhs \|^2_F = \sum_{ij} \target_{ij} \oplus [\lhs \circ \rhs]_{ij}\;.
        \label{eq:bmf}
    \end{equation}
\end{problem}
While beautiful in theory, this problem is NP-complete~\cite{Miettinen:2020:Recent}. 
Thus, we cannot solve this problem exactly for all but the smallest matrices.
In practice, we hence have to rely on approximations.
Here, we relax the Boolean constraints of Eq.~\eqref{eq:bmf} to allow non-negative, \emph{non-Boolean} `intermediate' factor matrices during the optimization, allowing us to use linear algebra rather than Boolean algebra. 
In other words, we solve the non-negative matrix factorization (NMF) problem~\cite{Paatero:1994:Positive}
\begin{equation}
    \| \target - \lhs\rhs \|^2_F \;, %
\end{equation}
subject to $\lhs \in \Real_+^{n \times k}$ and $\rhs \in \Real_+^{k \times m}$.
In contrast to the original BMF formulation, we can solve this problem efficiently, e.g., via a Gauss-Seidel scheme.
Although efficient, using plain NMF, however, disregards the Boolean structure of our matrices and produces factor matrices from a different domain, which are consequently hard to interpret and potentially very dense. 
To benefit from efficient optimization and still arrive at Boolean outputs, 
we allow real-valued intermediate solutions and regularize them towards becoming Boolean. 

\begin{figure}[!t]
    \begin{minipage}{.5\textwidth}
        \begin{figure}[H]
            \centering
            \includegraphics[height=5cm]{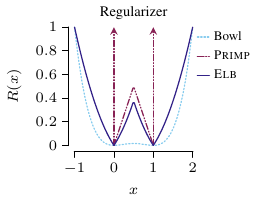}
        \end{figure}
    \end{minipage}
    \begin{minipage}[c]{.5\linewidth}
        \begin{algorithm2e}[H]
        \small
        \SetAlCapSty{}
        \DontPrintSemicolon
        \SetKwIF{If}{ElseIf}{Else}{if}{}{else if}{else}{end if}%
        \SetKwFor{While}{while}{}{end while}%
        \SetNlSty{tiny}{}{}%
        \KwIn{Matrix $\target \in \Bin^{n \times m}$, rank $k \in \Nat$}
        \KwOut{Factors $\lhs \in \Bin^{n \times k}$, $\rhs \in \Bin^{k \times m}$ }
        \BlankLine\BlankLine
        initialize $\lhs$ and $\rhs$ uniformly at random \; 
        \For{$t = 1,2,\dots$ \rm \textbf{until convergence}}{
            $\lhs \gets \operatorname{arg\ reduce}_{\lhs} \ell_{\kappa\lambda_t}(\target, \lhs, \rhs)$ \;
            $\rhs \gets \operatorname{arg\ reduce}_{\rhs} \ell_{\kappa\lambda_t}(\target, \lhs, \rhs)$ \;
        }
        \Return{$\lhs, \rhs$}
        \caption{\ourmethod} %
        \label{alg:ourmethod}
    \end{algorithm2e}
\end{minipage}

\caption{%
         On the left, we show the three regularizers: Bowl, \pimp, and \ourregularizer, for $\lambda = \kappa = 0.5$, and see that only our \ourregularizer regularizer penalizes non-Boolean values well.
         On the right, we show our method \ourmethod as pseudocode.}
         \label{fig:regularizer}
\end{figure}

To steer our optimization towards Boolean solutions, we penalize non-Boolean solutions using a regularizer.
This idea has been explored in prior work.
There exists the $l_1$-inspired \pimp regularizer~\cite{Hess:2017:PRIMPING}, which is 
\(-\kappa [-|1-2x| + 1]\) for values inside $[0, 1]$ and $\infty$ otherwise,
and the $l_2$-inspired bowl-shaped regularizer~\cite{Zhang:2007:Binary}, which is $\lambda (x^2 - x)^2/2$ everywhere on the real line.
Although both have been successfully applied to BMF, both also have undesirable properties: 
The \pimp regularizer penalizes well \emph{inside} the interval $[0, 1]$ but is non-differentiable on the outside,
while the bowl-shaped regularizer is differentiable and penalizes well \emph{outside} the interval $[0, 1]$ but is almost flat on the inside. 
Hence, both regularizers are problematic if used individually.
Combining them, however, yields a regularizer that penalizes non-Boolean values well across the full real line. 
To combine $l_1$- and $l_2$-regularization, we use the \emph{elastic-net regularizer},
\begin{equation*}
    r(x) = \kappa \|x\|_1 + \lambda \|x\|_2^2 \;,    
\end{equation*}
which, however, only penalizes \emph{non-zero} solutions~\cite{Zou:2005:Regularization}.
To penalize \emph{non-Boolean} solutions, we combine two elastic-net regularizers into our (almost \textrm{W}-shaped)
\textit{\ourregularizer regularizer}, 
\begin{equation}
    R(X) = \sum_{x \in X} \min \{ r(x), r(x - 1) \}\;,
    \label{eq:regularizer}
\end{equation}
where $X \in \{\lhs, \rhs\}$. 
In Fig.~\ref{fig:regularizer}, we show all three regularizers in the range of $[-1, 2]$, for $\lambda = \kappa = 0.5$.
We see that only the \ourregularizer regularizer penalizes non-Boolean solutions across the full spectrum, and
summarize our regularized relaxed BMF as follows.
\begin{problem}[Elastic Boolean Matrix Factorization]
    For a given target matrix $\target \in \Bin^{n \times m}$ and a given matrix rank $\Nat \ni k \leq \min\{n, m\}$, discover the factor matrices $\lhs \in \Real_+^{n \times k}$ and $\rhs \in \Real_+^{k \times m}$ that minimize 
    \begin{equation}
        \| \target - \lhs \rhs \|^2_F + R(\lhs) + R(\rhs) \; . %
        \label{eq:elbmf}
    \end{equation}
    \label{prob:elbmf}
\end{problem}

Although this is a relaxed problem, it is still non-convex, and therefore, we cannot solve it straightforwardly.
The problem, however, is suitable for the Gauss-Seidel optimization scheme. 
That is, we alternatingly fix one factor matrix to optimize the other. 
By doing so, we generate a sequence  
\begin{align}
    \lhs_{t+1} &\gets \argmin_{\lhs} \| \target - \lhs \rhs_{t}   \|^2_F + R(\lhs) \;,\\
    \rhs_{t+1} &\gets \argmin_{\rhs} \| \target - \lhs_{t+1} \rhs \|^2_F + R(\rhs) \;,
    \label{eq:sequence}
\end{align}
of simpler-to-solve sub-problems, until convergence.
Now, each sub-problem is again a sum of two $f(X) + R(X)$ functions, where $f$ is the loss $\| \cdot \|_F^2$, and $R(X)$ is the regularizer.
This allows us to follow a proximal gradient approach,
i.e., we use \emph{Proximal Alternating Linear Minimization} (PALM)~\cite{Bolte:2014:Proximal,Pock:2016:Inertial}.
In a nutshell, we minimize a sub-problem by following the gradient $\nabla f$ of $f$, to then use the proximal operator for $R$ to nudge its outcome towards a Boolean solution. 
That is, for the gradients $\nabla_\lhs f = \lhs\rhs\rhs^\Transpose - \target\rhs^\Transpose$ and 
$\nabla_{\rhs} f = \lhs^\Transpose\lhs\rhs - \lhs^\Transpose\target$, 
we compute the step
\begin{equation}
   \prox_R(X - \eta \nabla f) \;, 
   \label{eq:prox_step}
\end{equation}
where $\eta$ represents the step size, which we compute in terms of Lipschitz constant, rather than relying on a costly line-search~\cite{Bolte:2014:Proximal}. 
To further improve the convergence properties, we make use of an inertial term that linearly combines $X_{t}$ with $X_{t-1}$ before applying Eq.~\eqref{eq:prox_step} (see \cite{Pock:2016:Inertial} for a detailed description).
We now derive the proximal operator for the \ourregularizer regularizer, before discussing how we ensure that the factor matrices are Boolean, and summarizing our approach as an algorithm.
    
\subsection{Proximal Mapping}
To solve the sub-problem 
\begin{equation}
    \argmin_{X} \ell_{\kappa\lambda}(X)\ \text{for}\ \ell_{\kappa\lambda}(X) = f(X) + R(X) %
\end{equation} 
from Eq.~\eqref{eq:sequence} for $X \in \{\lhs,\rhs\}$,
we need a proximal operator~\cite{Parikh:2014:Proximal} that projects values towards a regularized point.
Formalized in Appendix~\ref{apx:prox}, in a nutshell, this operator is the solution to 
\begin{equation}
    \prox_{R}(X) = \argmin_Y \frac{1}{2} \| X - Y \|_2^2 + R(X) \;.
    \label{eq:prox_general}
\end{equation}

As Eq.~\eqref{eq:prox_general} is coordinate-wise solvable,
we can reduce this equation to a scalar proximal operator
\begin{equation}
    \prox_{\kappa\lambda}(x) = \argmin_{y \in \Real} \frac{1}{2} ( x - y )^2 + R(y) \;.
    \label{eq:prox_scalar}
\end{equation}
Although this scalar function may seem simple, due to $R$, it is non-convex, which---in general---rules out the existence of unique minima.
We can, however, exploit the W-shape of $R$. 
That is, from the perspective of a fixed $X_{ij}$, the regularizer is locally convex after a case distinction. 
Starting with the case of $X_{ij} = x,\ x \leq \frac{1}{2}$,
we can simplify Eq.~\eqref{eq:prox_scalar} to $\frac{1}{2}(x - y)^2 + \lambda'/2 y^2 + \kappa |y|$, letting $\lambda' = 2\lambda$.
Setting its derivative to $0$, we get to $y = x - \kappa \sign(y)$.
Asserting that a least-squares solution will always be of the same sign, we substitute the sign of $x$ with $\sign(y)$.
Repeating these steps analogously for $x > \nicefrac{1}{2}$, we obtain our proximal operator:
\begin{definition}[Proximal Operator]
    Given regularization coefficients $\kappa$ and $\lambda$, 
    our proximal operator for matrix $X$ is $\prox_R(X) = [\prox_{\kappa\lambda}(X_{ij})]_{ij}$, 
    where $\prox_{\kappa\lambda}$ is the scalar proximal operator
    \begin{equation}
        \prox_{\kappa\lambda}(x) \equiv 
            (1 + \lambda)^{-1} 
            \begin{cases}
                x - \kappa \sign(x)               & \textup{ if } x \leq \frac{1}{2} \\
                x - \kappa \sign(x - 1) + \lambda & \textup{ otherwise .}            \\
            \end{cases}
        \label{eq:elprox}
    \end{equation}
\end{definition}
Although not strictly necessary, to improve the empirical convergence rate, we would like to constrain our factor matrices to be non-negative.
As our proximal operator does not account for this, 
we impose non-negativity by using the \emph{alternating projection} procedure to combine the non-negativity proximal operator~\cite{Parikh:2014:Proximal} with Eq.~\eqref{eq:elprox} into $\max\{0, \prox_{\kappa\lambda}(x)\}$.

\subsection{Ensuring Boolean Factors}
Our proximal operator only nudges the factor matrices towards \emph{becoming} Boolean. 
We, however, want to ensure that our results \emph{are} Boolean. 
To this end, the state-of-the-art method \pimp relies heavily on post-processing, 
performing a very expensive joint two-dimensional grid search to guess the `best' pair of rounding thresholds, which are then used to produce Boolean matrices.
Although this tends to work in practice, it is an inefficient post-hoc procedure---%
and thus, it would be highly desirable to have Boolean factors already upon convergence.
To achieve this without rounding or clamping, we revisit our regularizer, which binarizes more strongly if we regularize more aggressively. 
Consequently, if we regularize too aggressively, we converge to a suboptimal solution,
and if we regularize too mildly, we do not binarize our solutions.
To prevent subpar solutions and still binarize our output, we start with a weak regularization and gradually increase its strength.

Considering Eq.~\eqref{eq:elprox}, we see that a higher regularization strength increases the distance over which our proximal operator projects.
Thus, if we set the $l_1$-distance controlling $\kappa$ too high, we will immediately leap to a Boolean factor matrix, 
which will terminate the algorithm and yield a suboptimal solution.
Regulating the $l_2$-distance controlling $\lambda$ is a less delicate matter. 
Hence, we gradually increase $\lambda$ to prevent a subpar solution and achieve a Boolean outcome, 
using a \emph{regularization rate} 
\begin{equation}
    \lambda_t = \lambda \cdot \nu_t\quad\text{for}\quad\nu_t \geq 0\quad\forall t \geq 0
\end{equation}
that gradually increases the proximal distance at a user-defined rate.
In case \ourmethod has stopped without convergence, 
we bridge the remaining integrality gap by projecting the outcome onto its closed Boolean counterpart, using our proximal operator (see Fig.~\ref{fig:convergence}).

We summarize the considerations laid out above as \ourmethod in Alg.~\ref{alg:ourmethod}. 
The computational complexity of \ourmethod is bounded by the complexity of computing the gradient, which is identical to the complexity of matrix multiplication. 
Therefore, for all practical purposes, 
\ourmethod is sub-cubic $\bigoh\left( n^{2.807} \right)$ using Strassen's algorithm.

\section{Related Work}\label{sec:related}

\emph{Matrix factorization} is a well-established family of methods, whose members, such as SVD, PCA, or NMF, are used everywhere in machine learning.
Almost all matrix factorization methods operate on real-valued matrices, however, while BMF operates under Boolean algebra.
\emph{Boolean Matrix Factorization} originated in the combinatorics community~\cite{Monson:1995:survey} and was later introduced to the data mining community~\cite{Miettinen:2008:discrete}, where many cover-based BMF algorithms were developed~\cite{Belohlavek:2010:Discovery,Belohlavek:2018:New,Miettinen:2008:discrete,Miettinen:2014:Mdl4bmf}.
In recent years, BMF has gained traction in the machine learning community, which tends to tackle the problem differently.
Here, \emph{relaxation-based approaches} that optimize for a relaxed but regularized BMF \cite{Hess:2017:PRIMPING,Hess:2017:Csalt,Zhang:2007:Binary} 
are related to our method, but they differ especially in their regularization.
Hess~et~al.~\cite{Hess:2017:PRIMPING} introduce a regularizer that is only partially differentiable, and they rely heavily on post-processing to force a Boolean solution, 
and Zhang~et~al.~\cite{Zhang:2007:Binary} regularize only weakly between $0$ and $1$.  
In contrast, our regularizer penalizes well across the full spectrum and yields a Boolean outcome upon convergence.
Building on a thresholding-based BMF formulation, Araujo~et~al.~\cite{Araujo:2016:FastStep} also consider relaxations to benefit from gradient-based optimization.
Other recent approaches build on \emph{probabilistic inference}.
Rukat~et~al.~\cite{Rukat:2017:interpretable,Rukat:2017:Bayesian,Rukat:2018:Probabilistic}, for example, combine Bayesian Modeling and sampling into their logical factor machine.
A similar direction is taken by Ravanbakhsh~et~al.~\cite{Ravanbakhsh:2016:Boolean}, who use graphical models and message passing,
and Liang~et~al.~\cite{Liang:2020:BEM}, who combine MAP-inference and sampling.
A different, \emph{geometry-based approach} lies in locating dense submatrices by ordering the data to exploit the consecutive-ones property~\cite{Wan:2020:Fast,Tatti:2019:Boolean}.
Since BMF is essentially solving a bipartite graph partitioning problem, it is also closely related to Bi-Clustering and Co-Clustering~\cite{Neumann:2018:Bipartite,Hess:2021:BROCCOLI}.
Neumann and Miettinen~\cite{Neumann:2020:Biclustering} use this relationship to efficiently solve BMF by means of a streaming algorithm.
Although there are many different approaches to BMF, its biggest challenge to date remains scalability~\cite{Miettinen:2020:Recent}.

\section{Experiments}\label{sec:experiments}

We implement \ourmethod in the Julia language and run experiments on 16 Cores of an AMD EPYC 7702 and a single NVIDIA A100 GPU, reporting wall-clock time.
We provide the source code, datasets, synthetic dataset generator, and other information needed for reproducibility.\!\footnote{Appendix~\ref{apx:reproducibility}; DOI: \ourdoiurl}
We compare \ourmethod against six methods: 
four dedicated BMF methods (\asso~\cite{Miettinen:2008:discrete}, \grecond~\cite{Belohlavek:2010:Discovery}, \LOM~\cite{Rukat:2017:Bayesian}, and \pimp~\cite{Hess:2017:PRIMPING}),
one streaming Bi-Clustering algorithm~\sofa~\cite{Neumann:2020:Biclustering},
one elastic-net-regularized NMF method leveraging proximal gradient descent (NMF~\cite{Paatero:1994:Positive,Lee:1999:Learning,Lee:2000:Algorithms}),
and one interpretable Boolean autoencoder (\binaps~\cite{Fischer:2021:Differentiable}). 
Since \nmf outputs non-negative factor matrices, rather Boolean matrices, we cannot compare against \nmf directly, so we clamp and round its solutions to the nearest Boolean outcome.
To fairly compare against \binaps, we task it with autoencoding the target matrix as a reconstruction, 
given the matrix ranks from our experiments as the number of latent dimensions.
We perform three sets of experiments.
First, we ascertain that \ourmethod works reliably on synthetic data.
Second, we verify that it generally performs well on real-world data.
And third, we illustrate that its outputs are semantically meaningful through an exploratory analysis of a biomedical dataset.

\subsection{Performance of \ourmethod on Synthetic Data}
In the following experiments, we ask four questions:
{(\textbf{1})} How does \ourmethod converge?; 
{(\textbf{2})} How well does \ourmethod recover the information in the target matrix?; %
{(\textbf{3})} How consistently does \ourmethod reconstruct low- or high-density target matrices?; and
{(\textbf{4})} Does \ourmethod estimate the underlying Boolean matrix rank correctly?
To answer these questions, we generate synthetic data with known ground truth as follows.
Starting with an all-zeros matrix, we randomly create rectangular, non-overlapping, consecutive areas of ones called \emph{tiles}, 
each spanning a randomly chosen number of consecutive rows and columns, thus inducing matrices with varying densities.
We then add noise by setting each cell to $1$, uniformly at random, with varying noise probabilities. 

\paragraph{How does \ourmethod converge?}
 
To study how our method converges to a Boolean solution, we quantify relevant properties of the sequence of intermediate solutions (cf. Eq.~\eqref{eq:sequence}). 
First, to understand how quickly and stably \ourmethod converges to a Boolean solution, 
we quantify the \emph{Boolean gap}, 
\[ \sum_{X\in \{U_t,V_t\}} |X|^{-1} \sum_{x\in X} \min\{ |x|, |x - 1|\}\;. \]
Second, to understand when we can safely round intermediate almost-Boolean solutions without losing information,
we calculate, for the reconstruction $\recons$ from \emph{rounded} intermediate factors,
the cumulative \emph{Hamming process} as the sum of fraction of bits that flip from iteration $t$ to iteration $t+1$, 
\[ |A|^{-1}\|\recons_{t} - \recons_{t+1} \|_1 \;,\]%
and the \emph{loss gap} as the difference between the relaxed loss and the loss from the \emph{rounded} $\recons$. 

\begin{figure*}[t] 
    \centering
    \includegraphics{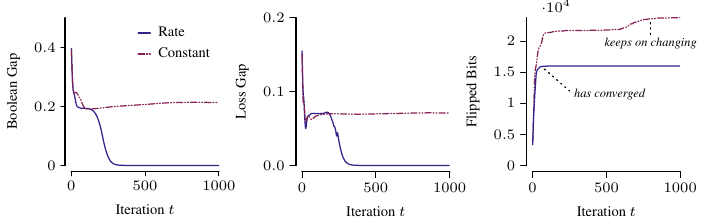}
    \caption{
        Our method converges quickly under a regularization rate.
        We report the progression over time of the
        \emph{Boolean gap}, 
        \emph{loss gap}, 
        and the \emph{Hamming process},
        for $1\,000$ iterations of \ourmethod on synthetic $400\times300$ matrices with $10\%$ noise and $5$ random tiles covering between $50$ and $100$ rows or columns, under a constant regularization of $\lambda_t = 1$ or a regularization rate of $\lambda_t = 1.05^t$.
    }
    \label{fig:convergence}
\end{figure*}

As shown in Fig.~\ref{fig:convergence}, we achieve an almost-Boolean solution \emph{without any rounding} after around $250$ epochs, continuing until we reach a Boolean outcome.
This is also the point at which the rounded intermediate solution and its relaxation are almost identical, 
as illustrated by the loss gap. 
Considering the Hamming process on the right, 
we observe that \ourmethod goes through an erratic bit-flipping-phase in the beginning, 
followed by only minor changes in each iteration until iteration $t = 100$.
Afterwards, \ourmethod has settled on a solution---under our regularization rate.
When using constant regularization instead, we continue to observe bit flips until the end of the experiment.
Under constant regularization, the Boolean gap hardly decreases over time.
Far from Boolean, the constant regularization thus also never closes the loss gap,
which is unsurprising, given that its factors are less regularized.
In other words, our regularization works well, and it allows us to safely binarize almost-converged factors that are $\epsilon$-far from being Boolean by means of, e.g., our proximal operator.

\newcommand{\recallgt}{$\text{recall}^*$}

\begin{figure*}[t] 
    \centering
    \includegraphics{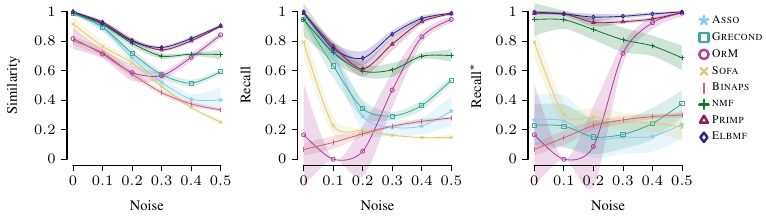}
    \caption{
        Overall, \ourmethod reconstructs the noisy synthetic data well and recovers the ground-truth tiles. 
        On synthetic data for additive noise levels increasing from $0\%$ to $50\%$, 
        we show mean as line and standard deviation as shade of \emph{similarity}, \emph{recall} w.r.t. the target matrix, and \emph{\recallgt}\ w.r.t. the noise-free ground-truth tiles, for \asso, \grecond, \LOM, \sofa, \binaps, \nmf, \pimp, and \ourmethod.   
    }
    \label{fig:noise}
\end{figure*}

\paragraph{How well does \ourmethod recover the information in the target matrix?}\label{sec:noise}
Having ensured that our method converges stably and quickly, we would like to assess whether it also converges to a high-quality factorization.  
To this end, we generate synthetic $40 \times 30$ matrices containing $5$ random tiles each spanning $5$ to $10$ rows and columns, under additive noise levels between $0\%$ (no noise) and $50\%$.
We then compute the fraction of \emph{ones} 
in target \target\ that is covered by the reconstruction $\recons = \lhs\circ\rhs$, i.e., the \emph{recall} (higher is better)
\begin{equation}
\|\target\|_1^{-1} \| \target \odot \recons \|_1\;.
\label{eq:recall}
\end{equation}
To ensure that we fit the \emph{signal} in the data, we additionally report the recall regarding the generating, noise-free ground-truth tiles $A^*$, denoted as \recallgt.
Finally, 
to rate the overall reconstruction quality including \emph{zeros},
we compute the \emph{Hamming similarity} (higher is better) between the target matrix and its reconstruction %
\[
|\target|^{-1} \|\target - \recons \|_1\;.
\]
We run each method on our synthetic datasets, targeting a matrix rank of $5$. 
To account for random fluctuations, we average over $10$ randomly drawn sets of $5$ ground-truth tiles per $10\%$ increment in noise probability. 
In Fig.~\ref{fig:noise}, we show similarity, recall, and \recallgt.
We observe that in the noiseless case ($0\%$), all methods except \binaps recover the $5$ ground-truth tiles with high accuracy,
but only \asso and \ourmethod do so with perfect recalls. 
Starting with as little as $10\%$ noise, both recalls of \asso, \grecond, \sofa, \nmf, and \binaps deteriorate quickly,
while the similarity and both recalls of \pimp and \ourmethod remain high.
In fact, \ourmethod and \pimp perform similarly across the board---%
which is highly encouraging, as unlike \pimp, \ourmethod does not require post-processing.
For \asso and \grecond, recall and similarity drop considerably, but they exhibit a slightly higher \recallgt.
This means that these methods are robust against noise, but they fail to recover the remaining information.
Starting low, \LOM's recalls increase jointly with the noise level, suggesting that clean data is problematic for \LOM.
Reporting the standard deviations as the shaded region, 
we see little variance across all similarities---except for \asso and \nmf in the highest noise regime.
The deviation of both recalls is, however, inconsistent for most methods, except for \binaps, \pimp, and \ourmethod. 
Overall, the performance characteristics of \ourmethod are among the most reliable.

\begin{figure*}[t] 
    \centering
    \includegraphics{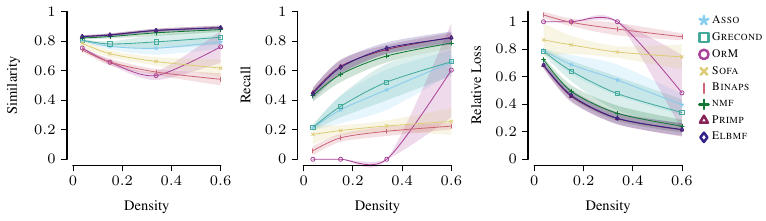}
    \caption{
        \ourmethod reconstructs noisy synthetic high-density and low-density matrices consistently well. 
        On synthetic data with fixed additive noise, and increasing density, we show 
        mean as line and standard deviation as shade of
        \emph{similarity}, \emph{recall}, and \emph{relative loss}, for \binaps, \asso, \grecond, \LOM, \sofa, \binaps, \nmf, \pimp, and \ourmethod.
    }
    \label{fig:density-high-noise}
\end{figure*}
\paragraph{How robust is \ourmethod regarding varying matrix densities?}\label{sec:densities}

To understand whether \ourmethod performs consistently well on low- or high-density matrices,
we generate synthetic matrices as before, this time using fixed noise of $20\%$, and varying the width and height of the ground-truth squared tiles from $3^2$ to $12^2$, resulting in densities between $0.0375$ to $0.6$, before noise. 

In Fig.~\ref{fig:density-high-noise}, we show the similarity, recall, and loss of \binaps, \asso, \grecond, \LOM, \sofa, \nmf, \pimp, and \ourmethod.
We can see that the increasing density affects the performance of all methods, however, it does not affect the performance of all methods \emph{equally}.
All methods---except \LOM---improve in similarity, recall, and loss.
With increasing density, \LOM gets worse at first, before its loss shrinks significantly, such that it finishes outperforming \sofa and \binaps.
From low to high density, \ourmethod is the best-performing method across the board in similarity, recall, and loss.

\paragraph{Does \ourmethod estimate the underlying Boolean matrix rank correctly?}
When a target rank is known, we can immediately apply \ourmethod to factorize the data. 
In the real world, however, the target rank might be unknown.
In this case, we need to \emph{estimate} an appropriate choice from the data, and we use synthetic data to ensure that \ourmethod does so correctly.

Since a higher matrix rank usually also means a better fit, 
selecting the best rank according to recall, loss, or similarity leads to overfitting---%
unless we properly account for the growth in model complexity.
There are many \emph{model selection criteria} that penalize complex models, such as 
AIC~\cite{Akaike:1974:New}, 
Bai \& Ng's criteria~\cite{Bai:2002:Determining,Alessi:2010:Improved}, 
Nuclear-norm regularizing~\cite{Jaggi:2010:Simple,Gunasekar:2017:Implicit}, 
the information-theoretic Minimal Description Length principle (MDL)~\cite{Grunwald:2007:Minimum},
or (Decomposed) Normalized Maximum Likelihood~\cite{Ito:2016:Rank,Wu:2017:Decomposed}.
Following common practice, and motivated by its practical performance in preliminary experiments, we choose MDL.
That is, we select the \emph{minimizer} of the sum of the log binomial \( l(X) = \log \binom{\abs{X}}{\|X\|_1} \)
of the error matrix and the rows and columns of our factorization (assumed to be i.i.d.)~\cite{Miettinen:2014:Mdl4bmf} %
\[
\textstyle    
l(\target \oplus [\lhs \circ \rhs]) + \sum_{i \in [k]} l(\lhs^\Transpose_i) + l(\rhs_i) + k\log(n \cdot m) \;.
\]

\begin{figure*}[t] 
    \centering
    \includegraphics{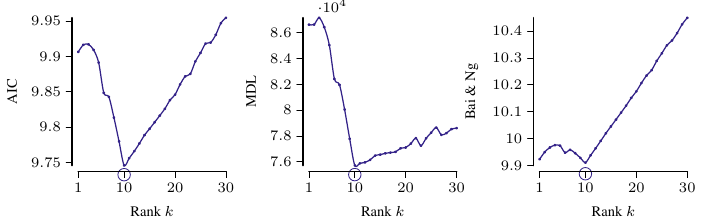}
    \caption{
        Using AIC, MDL, or Bai \& Ng's first information criteria, \ourmethod correctly detects the simulated $400\times 300$ matrix of rank $10$ to which we applied $10\%$ additive noise.
    }
    \label{fig:model-selection}
\end{figure*}

To validate whether \ourmethod recovers the correct rank, 
we synthetically generate a $400\times 300$ matrix of ground-truth rank $10$ with $10\%$ additive noise.
In Fig.~\ref{fig:model-selection}, we show AIC, MDL, and Bai \& Ng's first criterion for each rank up to $30$, 
finding that our method precisely discovers the right rank. 

\subsection{Performance of \ourmethod on Real-World Data}
Having ascertained that \ourmethod works well on synthetic data, we turn to its performance in the real world.
Here, we use nine publicly available datasets\footnote{%
    \smaller
    \href{https://grand.networkmedicine.org}{\textsc{grand.networkmedicine.org}}
    \xspace\href{https://string-db.org}{\textsc{string-db.org}}
    \xspace\href{https://cancer.gov/tcga}{\textsc{cancer.gov/tcga}}
    \xspace\href{https://www.internationalgenome.org/}{\textsc{internationalgenome.org}}
    \xspace\href{https://patentsview.org/download/data-download-tables}{\textsc{patentsview.org}}
    \xspace\href{https://grouplens.org/datasets/movielens/}{\textsc{grouplens.org/datasets/movielens} }
    \xspace\href{https://www.kaggle.com/datasets/netflix-inc/netflix-prize-data}{\textsc{kaggle.com/datasets/netflix-inc/netflix-prize-data}}
}
from different domains.
To cover the \emph{biomedical domain}, 
we extract the network containing empirical evidence of protein-protein interactions in \emph{Homo sapiens} from the \textsc{String} database.
From the \textsc{Grand} repository, 
we take the gene regulatory networks sampled from \emph{Glioblastoma (GBM)} and \emph{Lower Grade Glioma (LGG)} brain cancer tissues, 
as well as from non-cancerous \emph{Cerebellum} tissue.
The \emph{TCGA} dataset contains binarized gene expressions from cancer patients,
and we further obtain the single nucleotide polymorphism (SNP) mutation data from the \emph{$1$k Genomes} project,
following processing steps from the authors of \binaps~\cite{Fischer:2021:Differentiable}.
In the \emph{entertainment domain}, we use the user-movie datasets \emph{Movielens} and \emph{Netflix}, binarizing the original $5$-star-scale ratings by setting only reviews with more than $3.5$ stars to $1$.
Finally, as data from the \emph{innovation domain}, 
we derive a directed citation network between patent groups from patent citation and classification data provided by \emph{PatentsView}.
For each dataset with a given number of groups, such as cancer types or movie genres, we set the matrix rank $k$ to 33 (TCGA), 28 (Genomes), 136 (Patents), 20 (Movielens), and 20 (Netflix).
When the number of subgroups is unknown, we estimate the rank that minimizes MDL using \ourmethod, resulting in  
100 (GBM), 32 (LGG), 100 (String), and 450 (Cerebellum). 

As we can achieve a high similarity with an all-zeros reconstruction (in the case of sparse data), or a perfect recall with an all-ones reconstruction,
we also report the \emph{relative loss} (lower is better), 
\[
    \|\target\|_1\|\target - \lhs \circ \rhs\|_1\;,
\]
between the target matrices and their reconstructions.

\begin{figure*}[tb] 
    \centering
    \includegraphics{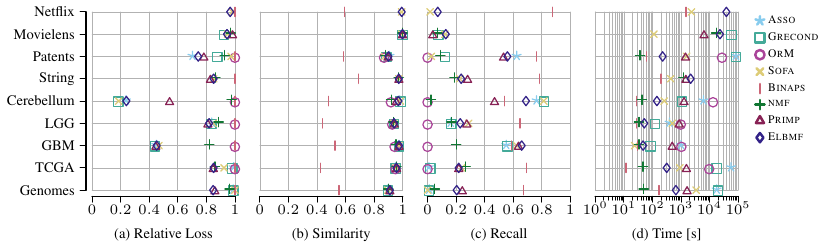}
    \caption{
        Our method factorizes real-world data with high similarity and recall, as well as low relative loss and runtime.
        We report \emph{relative loss}, \emph{similarity}, \emph{recall}, and \emph{runtime} of $9$ real-world matrices and their reconstructions from \asso, \grecond, \LOM, \sofa, \binaps, \nmf, \pimp, and \ourmethod. 
    }
    \label{fig:realworld}
\end{figure*}

We show relative loss, similarity, recall, and runtime of \asso, \grecond, \sofa, \LOM, \binaps, \nmf, \pimp, and \ourmethod, applied to all real-world datasets, 
targeting a given matrix rank, in Fig.~\ref{fig:realworld}.
The cover-based \grecond and \asso show comparable loss, similarity, and recall.
Both perform better on smaller matrices (\emph{LGG}, \emph{GBM}, or \emph{Cerebellum}) and struggle with complex matrices (e.g., \emph{TCGA}, \emph{Genomes}). 
On the complex matrices (e.g., on \emph{TCGA}, \emph{String}, or \emph{Movielens}), 
although always outperformed by \ourmethod, we see that the \emph{rounded} \nmf reconstructions are surprisingly good, occasionally surpassing dedicated BMF methods, such as \asso, \grecond, \LOM, and \sofa.
Across the board, \grecond, \asso, \LOM, \sofa, \binaps, and \nmf frequently result in considerably higher loss than \ourmethod.
Compared to the close competitor \pimp, our method \ourmethod always results in lower reconstruction loss.
We observe the largest gap between the two on the Cerebellum dataset, where \pimp's grid search procedure fails to find suitable thresholds.
This is an impressive result because unlike \pimp, \ourmethod does not necessitate heavy post-processing.

In Fig.~\ref{fig:realworld}b, we see that all methods except \binaps result in a high similarity, which implies they are sparsity-inducing.
As \binaps overfits and densely reconstructs sparse inputs, it surpasses sparsity-inducing methods in recall.
Considering non-overfitting methods, however, \ourmethod is among the best performing in terms or recall, 
often outperforming \pimp, while under significantly stronger regularization.
When \pimp has a higher recall (e.g., \emph{Genomes}), this often comes with a higher loss than \ourmethod.
Noticeable is \sofa's improvement relative to its performance on synthetic data.
We see in Fig.~\ref{fig:realworld}d (log scale) that in most cases \asso, \grecond, \LOM, \sofa, and \pimp are slower than \ourmethod.
Although \nmf is less constrained than \ourmethod, both are almost on par when it comes to runtime.
Degraded by post-processing, our closed competitor \pimp is almost always much slower than \ourmethod and struggles with \emph{Netflix}.
Only \binaps and \ourmethod finished \emph{Netflix}, however, only \ourmethod did so at a reasonable loss, considering the given target rank.

\subsection{Exploratory Analysis of Gene Expression Data with \ourmethod}
Knowing that \ourmethod performs well quantitatively, we ask whether its outputs are also interpretable.
To this end, we take a closer look at the \emph{TCGA} data, 
which contains the expression levels of $20\ 530$ genes from $10\ 459$ patients, 
who are labeled with $33$ cancer types. 
Since we are interested in retaining high gene expression levels only, we set expression levels to one if their $z$-scores fall into the top $5\%$ quantile, and to zero otherwise~\cite{Liang:2020:BEM}.
We run \ourmethod on this dataset, targeting a rank of $33$.  

\begin{figure*}[t] 
    \centering
    \subfloat[TCGA]{\includegraphics{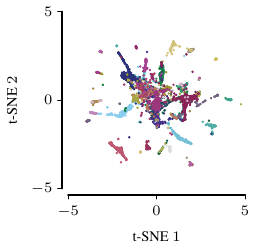}}
    \subfloat[Reconstructed TCGA]{\includegraphics{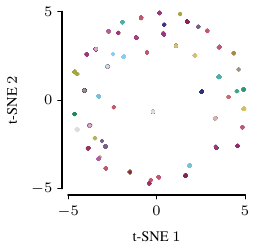}}  
    \subfloat[Normalized Mutual Information]{\includegraphics{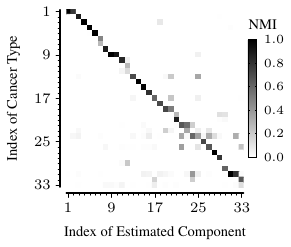}}   
    \caption{
        \ourmethod discovers the hidden structure in gene expression data.
        We show the two-dimensional t-SNE embedding of the TCGA dataset (a) and the embedding of its reconstruction from \ourmethod (b), 
        where each point corresponds to one of the $10\ 459$ patients, colored by cancer type.
        While the cancer types are hard to differentiate in the embedding of the original dataset (a), 
        they are separated into easily distinguishable clusters in the embedding of our reconstruction (b).
        In (c), we show the normalized mutual information between estimated groups and cancer types, and observe that there is an almost $1$-to-$1$ correspondence between our estimated groups and the cancer types. 
    }
    \label{fig:tcga}
\end{figure*}

To learn whether our method groups patients meaningfully, we visualize the target matrix and its reconstruction.
As the target matrix is high-dimensional, we embed both the target matrix and the reconstruction into a two-dimensional space using t-SNE~\cite{VanderMaaten:2008:Visualizing} as illustrated in Fig.~\ref{fig:tcga},
where each color corresponds to one cancer type.
In Fig.~\ref{fig:tcga}a, we show that when embedding the target matrix directly, 
the cancer types are highly overlapping and hard to distinguish without the color coding.
In contrast, when embedding our reconstruction, depicted in Fig.~\ref{fig:tcga}b, we see a clean segmentation into 
clusters that predominantly contain a single cancer type.

To better understand these results, we quantify the association between our $33$ estimated components and the ground-truth cancer types by computing the \emph{normalized mutual information} matrix.
This matrix is noticeably sparse (Fig.~\ref{fig:tcga}c), leading to a clean segmentation.
Upon closer inspection with \textsc{Enrichr}~\cite{Chen:2013:Enrichr}, the associations we discover turn out to be biologically meaningful.
For example, we find that \ourmethod associates a set of $356$ genes to patients with \emph{thyroid carcinoma}.  
This component is associated with \emph{thyroid hormone generation} and \emph{thyroid gland development}, 
and \emph{statistically significantly} so---%
even under a strict False Discovery Control, with $p$-values as low as $2.574\times10^{-8}$ and $6.530\times10^{-6}$.

\section{Conclusion}\label{sec:conclusion}

We introduced \ourmethod to efficiently factorize Boolean matrices using an elegant and simple algorithm that, unlike its closest competitors, does not rely on heavy post-processing.
\ourmethod considers a novel relaxed BMF problem, which allows intermediate solutions to be real-valued. 
It solves this problem by leveraging an efficiently computable proximal operator, derived from the innovative \ourregularizer regularizer, and using a regularization rate to obtain Boolean factors upon convergence.
Experimentally, we have shown that \ourmethod works well in practice. 
It operates reliably on synthetic data, 
outperforms the discrete state of the art, 
is at least as good as the best relaxations on real data,
and yields interpretable results even in difficult domains---%
without relying on post-processing.

\textbf{Limitations}\quad Although \ourmethod works overall, it has two bottlenecks.
First, by randomly initializing factors, we start with highly dense matrices, thus prohibiting efficient sparse matrix operations.
This is not ideal for sparse datasets that are too large to fit into memory, 
and future research on sparse initialization of \emph{Boolean} factors will benefit not only \ourmethod but also many other methods. 
Second, the larger the datasets, the higher the cost of computing gradients,
and future work might adapt stochastic gradient methods for \ourmethod 
to mitigate this problem.

\textbf{Broader Impact}\quad \ourmethod is a method for factorizing Boolean matrices, which permits interpretable, rather than black-box data analysis.
As such, it can help make explicit any biases that may be present in the data.
Just like any other method for interpretable data analysis, the insights \ourmethod yields can be used for good purposes (e.g., insight into cancer-causing mutations) or for bad purposes, and the ultimate quality of the outcome also depends on the behavior of the user.
In our experiments, we focus on beneficial applications, and only consider anonymized open data.

\clearpage

\renewcommand{\acksection}{\section*{Acknowledgments}}
\begin{ack}
We thank Stefan Neumann for providing us with the tuned parametrization of \sofa~\cite{Neumann:2020:Biclustering}.
\end{ack}

\bibliographystyle{abbrvnat}

\clearpage
\appendix

\section*{Appendix}\label{sec:appendix}

We include the following supplementary materials in this Appendix.

In Appendix~\ref{apx:prox}, we formally derive our proximal operator from our \ourregularizer regularizer, resulting in the proximal operator Eq.~\eqref{eq:elprox} as well as an alternative proximal operator Eq.~\eqref{eq:elprox_alt}, not used in our experiments.

In Appendix~\ref{apx:alg}, we provide a more detailed version of Algorithm~\ref{alg:ourmethod}.

In Appendix~\ref{apx:reproducibility}, we supply details regarding reproducibility, such as hyper-parameters, data-preprocessing, implementation details, and datasets. 

In Appendix~\ref{apx:synth}, we share additional experiments on synthetic data under varying levels of additive noise, this time allowing for overlapping tiles.

In Appendix~\ref{apx:densities}, we provide additional experiments on synthetic data under a low additive noise and varying densities. 

\section{Derivation of the Proximal Operator}\label{apx:prox}
In this section, we derive our proximal operator (given in Eq.~\eqref{eq:elprox}, repeated below) 
\begin{equation}
    \prox_{\kappa\lambda}(x) \equiv 
        (1 + \lambda)^{-1} 
        \begin{cases}
            x - \kappa \sign(x)               & \textup{ if } x \leq \frac{1}{2} \\
            x - \kappa \sign(x - 1) + \lambda & \textup{ otherwise ,}            \\
        \end{cases}
\end{equation}
from the \ourregularizer (Eq.~\eqref{eq:regularizer}, repeated below)
\[ R(X) = \sum_{x \in X} \min \{ r(x), r(x - 1) \} \;.\]
Starting with the definition~\cite{Parikh:2014:Proximal} of the general proximal operator 
\[
    \argmin_Y \frac{1}{2} \| X - Y \|_2^2 + R(X) \;,
\]
we observe that this proximal operator is coordinate-wise solvable.
This allows us to derive a \emph{scalar} proximal operator, which we then apply to each value in the matrix independently. 
Substituting the regularizer $R$ with its scalar version, we obtain a scalar proximal operator (Eq.~\eqref{eq:prox_scalar}) 
\[     
    \prox_{\kappa\lambda}(x) = \argmin_{y \in \Real} \frac{1}{2} ( x - y )^2 +\min \{ r(y), r(y - 1) \} \;,
\] 
which is non-convex, has no unique minima, and, therefore, is not straightforwardly solvable.
We can, however, separate this function into two \emph{locally convex} V-shaped functions, which are straightforwardly solvable. 
By asserting that its least-squares solution is either
less than $\nicefrac{1}{2}$ (if $x \leq \nicefrac{1}{2}$), 
or greater than $\nicefrac{1}{2}$ (if $x > \nicefrac{1}{2}$),
we can address each case independently, and merge the outcome into a single piecewise proximal operator. 
\paragraph{Case \Rom{1}} 

In the first case, we address the operator for $x \leq \nicefrac{1}{2}$.
For this, we start by simplifying our scalar proximal operator Eq.~\eqref{eq:prox_scalar}, by substituting $r(y)$ with its definition, and get
\[ 
    \prox_{\kappa\lambda}^{\leq\nicefrac{1}{2}}(x) = \frac{1}{2} ( y - x )^2 + \frac{\lambda'}{2} y^2 + \kappa |y|\;,
\]
for $\lambda'/2 = \lambda$.
Then, we take its partial derivative for $y$
\[
    \frac{\partial}{\partial y} \prox_{\kappa\lambda}^{\leq\nicefrac{1}{2}}(x) = (y - x) + \lambda' y + \kappa \sign(y)\;,
\]
which we set to zero, obtaining
\begin{align*}
                        0 = (y - x) + \lambda'y + \kappa \sign(y) 
    \Leftrightarrow\;   0 = y(1 + \lambda') - x + \kappa \sign(y) \;.\\
\end{align*}
By asserting that we can obtain a better least-squares solution if $y$ has the same sign as $x$, 
we can substitute the sign of $x$ with $\sign(y)$, and get
\begin{align*}
                     0 = y(1 + \lambda') - x + \kappa \sign(x) 
    \Leftrightarrow\;  y = (1 + \lambda')^{-1}[x - \kappa \sign(x)] \;, \\
\end{align*}
which concludes the first case.

\paragraph{Case \Rom{2}} 
Analogously, we now repeat the steps from above for $x > \nicefrac{1}{2}$.
Again, we start by simplifying Eq.~\eqref{eq:prox_scalar}, substituting $r(y-1)$
\[ 
    \prox_{\kappa\lambda}^{>\nicefrac{1}{2}}(x) = \frac{1}{2} ( y - x )^2 + \frac{\lambda'}{2} (y-1)^2 + \kappa |y-1|\;,
\]
for $\lambda'/2 = \lambda$.
By taking its partial derivative for $y$
\[
    \frac{\partial}{\partial y} \prox_{\kappa\lambda}^{>\nicefrac{1}{2}}(x) = (y - x) + \lambda' (y-1) + \kappa \sign(y-1)\;,
\]
and setting it to zero, we obtain
\begin{align*}
                     0 = (y - x) + \lambda'(y-1) + \kappa \sign(y-1)
    \Leftrightarrow\ 0 = (1+\lambda')y - \lambda' - x + \kappa \sign(y-1) \;. \\
\end{align*}
Then, asserting that the least-squares solution does not get worse by using the same sign for $y-1$ and $x-1$, we can substitute the sign of $x-1$ with $\sign(y-1)$, and get
\begin{align*}
                      0 = y(1 + \lambda') - x - \lambda' + \kappa \sign(x-1)
    \Leftrightarrow\; y = (1 + \lambda')^{-1}[x - \kappa \sign(x-1) + \lambda'] \;,  \\
\end{align*}
which concludes the $x > \nicefrac{1}{2}$ case.

\paragraph{Combining Case \Rom{1} \& Case \Rom{2}}
Combining the cases above yields our piecewise proximal operator (see Eq.~\eqref{eq:elprox})
\begin{equation}
    \prox_{\kappa\lambda}(x) \equiv 
        (1 + \lambda)^{-1} 
        \begin{cases}
            x - \kappa \sign(x)               & \textup{ if } x \leq \frac{1}{2} \\
            x - \kappa \sign(x - 1) + \lambda & \textup{ otherwise .}            \\
        \end{cases}
\end{equation}

\paragraph{Alternative Proximal Operator}
Considering Eq.~\eqref{eq:prox_scalar}, we notice that the term $y - x$ is squared, which means that there are multiple solutions to this equation.
We derive the alternative operator analogously to the steps above, however, by switching the positions of $y$ and $x$ in $f$.
\begin{equation}
    \prox_{\kappa\lambda}^{\textrm{alt.}}(x) \equiv 
        (\lambda - 1)^{-1} 
        \begin{cases}
            -x - \kappa \sign(x)               & \textup{ if } x \leq \frac{1}{2} \\
            -x - \kappa \sign(x - 1) + \lambda & \textup{ otherwise .}            \\
        \end{cases}
        \label{eq:elprox_alt}
\end{equation}
Since this operator is denominated by $\lambda' - 1$, we need to ensure that $\lambda' \not= 1$.
Because our original proximal operator in Eq.~\eqref{eq:elprox} is denominated by $\lambda' + 1$, 
and since $\lambda'$ is usually positive, 
we are not required to take extra precautions.
Since this is more convenient, we select Eq.~\eqref{eq:elprox} as our proximal operator, rather than taking extra precautions when using Eq.~\eqref{eq:elprox_alt}.

\clearpage
\section{Extended pseudocode for \ourmethod}\label{apx:alg}

Extending Alg.~\ref{alg:ourmethod}, we provide the more detailed version of \ourmethod as pseudocode in Alg.~\ref{alg:long}.

\begin{algorithm2e}%
    \small
    \SetAlCapSty{}
    \DontPrintSemicolon
    \SetKwIF{If}{ElseIf}{Else}{if}{}{else if}{else}{end if}%
    \SetKwFor{While}{while}{}{end while}%
    \SetNlSty{tiny}{}{}%
    \KwIn{\begin{tabular}[c]{ll}
            Target Matrix                  & $\target \in \Bin^{n \times m}$\\
            Rank                           & $k \in \Nat$,\\
            $l_1$ Regularizer Coefficients & $\kappa \in \Real$,\\
            $l_2$ Regularizer Coefficients & $\lambda \in \Real$,\\
            Regularization Rate            & $\nu_t \in \Nat \to \Real$,\\
            \textbf{optional} Inertial Parameter  & $\beta \in \Real_+$ 
        \end{tabular}%
    }
    \KwOut{Factors $\lhs \in \Bin^{n \times k}$, $\rhs \in \Bin^{k \times m}$ }
    \BlankLine
    $\lhs_0 = \lhs_1 \gets \operatorname{rand}(n, k)$ \;
    $\rhs_0 = \rhs_1 \gets \operatorname{rand}(k, m)$ \;
    \BlankLine
    \For{$t = 1,2,\dots$ \rm \textbf{until convergence}}{
        $\lambda_t \gets \lambda \cdot \nu_t$\;
        \BlankLine
        \BlankLine
        $\lhs \gets \lhs_{t-1} + \beta (\lhs_{t-1} - \lhs_{t-2})$\;
        $\nabla_\lhs f = \lhs\rhs\rhs^\Transpose - \target\rhs^\Transpose$ \;
        $L \gets \|\rhs\rhs^\Transpose\|_2$\;
        $\lhs \gets \prox_{\kappa L^{-1},\lambda_t L^{-1}} \left( \lhs - L^{-1} \nabla_\lhs f \right)$\;
        $\lhs_t \gets \lhs$ \;
        \BlankLine
        \BlankLine
        $\rhs \gets \rhs_{t-1} + \beta (\rhs_{t-1} - \rhs_{t-2})$\;
        $\nabla_{\rhs} f = \lhs^\Transpose\lhs\rhs - \lhs^\Transpose\target$ \;
        $L \gets \|\lhs^\Transpose\lhs\|_2$\;
        $\rhs \gets \prox_{\kappa L^{-1},\lambda_t L^{-1}} \left( \rhs - L^{-1} \nabla_\rhs f \right)$\;
        $\rhs_t \gets \rhs$ \;
    }
    \BlankLine
    \If{\rm $\lhs$ or $\rhs$ not Boolean\qquad \emph{(i.e., if the above was aborted early (cf. Fig.~\ref{fig:convergence}))}}{ 
        let $\lambda' \in \Real$ be huge \;
        $\lhs \gets \lfloor \prox_{0.5,\lambda'}(\lhs) \rceil $ \;
        $\rhs \gets \lfloor \prox_{0.5,\lambda'}(\rhs) \rceil $ \;
    }
    \BlankLine
    \Return{$\lhs$, $\rhs$}
    \caption{Long version of \ourmethod in terms of iPALM~\cite{Pock:2016:Inertial}} %
    \label{alg:long}
\end{algorithm2e}

\section{Reproducibility}\label{apx:reproducibility}

\begin{table}[b]
    \centering
    \caption{%
        Our datasets are from different domains and cover a wide-range of dimensionalties. 
        We provide an overview over the real-world datasets involved in this study, listing their dimensionalities, densities, and selected target matrix ranks $k$ (number of components) used in our experiments.}
    
    \sisetup{round-mode=places, round-precision=4, table-format=3.4, scientific-notation=fixed, table-omit-exponent}
    \npthousandsep{\,}

    \begin{tabular}{lrN{7}{0}N{7}{0}l}
        \toprule
        Dataset          &Rank & Rows    & Columns   & Density                  \\
        \midrule
        Genomes          & 28& 2504    & 226623   & \num{0.10432025086095684}   \\
        String           &100& 19385   & 19385    & \num{0.03177006331327474}   \\
        GBM              &100& 650     & 10701    & \num{0.05664790494058787}   \\
        LGG              & 32& 644     & 29374    & \num{0.07294066202121537}   \\
        Cerebellum       &450& 644     & 30243    & \num{0.08230214147393689}   \\
        TCGA             & 33& 10459   & 20530    & \num{0.05006842528059488}   \\
        Movielens 10M    & 20& 71567   & 65133    & \num{0.0010738646228571796} \\
        Netflix          & 20& 17770   & 480189   & \num{0.006670510562061761}  \\
        Patents          &136& 10499   & 10511    & \num{0.1305279999620135}    \\
        \bottomrule
    \end{tabular}
    \label{tab:datasets}
\end{table}
Here, we explain 
{(\textbf{1})} how to obtain the datasets,
{(\textbf{2})} how we binarized them,
{(\textbf{3})} how to obtain the source code, and
{(\textbf{4})} how we parameterized the algorithms, 
starting with the dataset descriptions.

Broadly speaking, our datasets cover three domains: the \emph{biomedical domain}, the \emph{entertainment domain}, and the \emph{innovation domain}.
To cover the \emph{biomedical domain}, 
we extract the network containing empirical evidence of protein-protein interactions in \emph{homo sapiens} from the \textsc{String} database.
For this, we remove all interactions in the protein-protein network, for which the empirical-evidence sub-score of the \textsc{String} database is zero, 
retaining only the protein-protein interactions discovered experimentally.
From the \textsc{Grand} repository, 
we take the gene regulatory networks sampled from \emph{Glioblastoma (GBM)} and \emph{Lower Grade Glioma (LGG)} brain cancer tissues, 
as well as from non-cancerous \emph{Cerebellum} tissue.
These networks all stem from a method, \textsc{Panda}, which extracts gene-regulatory networks from data.
The weights are between $-20$ and $20$, where $-20$ corresponds to a low interaction likelihood, and $20$ corresponds to a high interaction likelihood.
To binarize, we set everything to zero except the weights whose $z$-scores are in the upper $5\%$ quantile, retaining the information about likely interactions.
The \emph{TCGA} dataset contains gene expressions from cancer patients.
To binarize these logarithmic expression rates ($\log_{10}(x + 1)$), we again only set those weights to one whose $z$-scores lie in the upper $5\%$ quantile~\cite{Liang:2020:BEM}, retaining high gene expressions.
We further obtain the single nucleotide polymorphism (SNP) mutation data from the \emph{$1$k Genomes} project,
and follow the data retrieval steps from the authors of \binaps~\cite{Fischer:2021:Differentiable}, 
which immediately produces a binary dataset.
In the \emph{entertainment domain}, we use the user-movie datasets \emph{Movielens} and \emph{Netflix}.
Since we are only interested in recommending good movies, we binarize the original $5$-star-scaled ratings, by setting reviews with more than $3.5$ stars to one, and everything else to zero.
Finally, as data from the \emph{innovation domain}, 
we derive a directed citation network between patent groups from patent citation and classification data provided by \emph{PatentsView}.
We binarize this weighted network simply by setting every non-zero weight to one, retaining all edges in the network.
For each of these publicly\footnote{%
\smaller
\href{https://grand.networkmedicine.org}{\textsc{grand.networkmedicine.org}}
\xspace\href{https://string-db.org}{\textsc{string-db.org}}
\xspace\href{https://cancer.gov/tcga}{\textsc{cancer.gov/tcga}}
\xspace\href{https://www.internationalgenome.org/}{\textsc{internationalgenome.org}}
\xspace\href{https://patentsview.org/download/data-download-tables}{\textsc{patentsview.org}}
\xspace\href{https://grouplens.org/datasets/movielens/}{\textsc{grouplens.org/datasets/movielens}}
\xspace\href{https://www.kaggle.com/datasets/netflix-inc/netflix-prize-data}{\textsc{kaggle.com/datasets/netflix-inc/netflix-prize-data}}
}
available dataset, we give its dimensionality, density, and the matrix rank used in our experiments in Table~\ref{tab:datasets}.

In our experiments in Sec.~\ref{sec:experiments}, we compare \ourmethod against six methods: 
four dedicated BMF methods (\asso~\cite{Miettinen:2008:discrete}, \grecond~\cite{Belohlavek:2010:Discovery}, \LOM~\cite{Rukat:2017:Bayesian}, and \pimp~\cite{Hess:2017:PRIMPING}),
one streaming Bi-Clustering algorithm~\sofa~\cite{Neumann:2020:Biclustering},
one elastic-net-regularized NMF method leveraging proximal gradient descent (NMF~\cite{Paatero:1994:Positive,Lee:1999:Learning,Lee:2000:Algorithms}),
and one interpretable Boolean autoencoder (\binaps~\cite{Fischer:2021:Differentiable}).
The code for \asso, \grecond, \pimp, \sofa, \LOM, and \binaps was written by their respective authors and is publicly available.
We implement \nmf and \ourmethod in the Julia programming language and provide their source code for reproducibility.\!\footnote{%
\smaller            
\href{https://cs.uef.fi/~pauli/basso/}{\textsc{cs.uef.fi/\textasciitilde{pauli}/basso}}
\quad\href{https://github.com/martin-trnecka/matrix-factorization-algorithms/}{\textsc{github.com/martin-trnecka/matrix-factorization-algorithms}}\\
\quad\href{https://bitbucket.org/np84/paltiling/src/master/}{\textsc{bitbucket.org/np84/paltiling}}
\quad\href{https://cs.uef.fi/~pauli/bmf/sofa/}{\textsc{cs.uef.fi/\textasciitilde{pauli}/bmf/sofa}}
\quad\href{http://eda.mmci.uni-saarland.de/prj/binaps/}{\textsc{eda.mmci.uni-saarland.de/prj/binaps}}\\
\quad\href{https://github.com/TammoR/LogicalFactorisationMachines}{\textsc{github.com/TammoR/LogicalFactorisationMachines}}
\quad{\textsc{\oururl}}
}
On \emph{TCGA}, \emph{Genomes}, \emph{Movielens}, \emph{Netflix}, and \emph{Patents}, we set the $l_2$-regularizer $\lambda = 0.001$, the $l_1$-regularizer $\kappa = 0.005$, and the regularization rate to $\nu_t = 1.0033^t$.
On \emph{GBM}, \emph{LGG}, and \emph{Cerebellum}, we set the $l_2$-regularizer $\lambda = 0.001$, the $l_1$-regularizer $\kappa = 0.001$, and the regularization rate to $\nu_t = 1.0015^t$.
We run \nmf, \ourmethod, \pimp for at most $1\ 500$ epochs on each dataset.
In the case that \ourmethod reaches its maximum number of iterations without convergence, 
we bridge the remaining integrality gap simply by applying our proximal operator (see Fig.~\ref{fig:convergence}).
To obtain a good reconstruction for \pimp, we use a grid-width of $0.01$.
To obtain a binary solution from \nmf, we first clamp and then round its factor matrices upon convergence.

We set \asso's threshold, gain for covering, and penalty for over-covering each to $1$.
To achieve a better performance with \asso, 
we parallelize \asso on $16$ CPU cores.
Further, because \asso's runtime scales with the number of columns, 
we reconstruct the \emph{transposed} target 
whenever it has more columns than rows (see Table~\ref{tab:datasets}).
For example, transposing \emph{GBM}, \emph{LGG}, \emph{Cerebellum}, and \emph{Genomes} is particularly beneficial for \asso, 
as these datasets have orders-of-magnitude more columns than rows.

\section{Additional Results on the Performance of \ourmethod on Synthetic Data}\label{apx:synth}

\begin{figure*}[t] 
    \centering
    \includegraphics{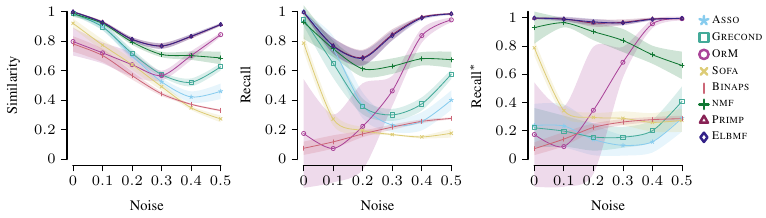}
    \caption{
        Overall, \ourmethod reconstructs the noisy synthetic data well and recovers the ground-truth tiles which are \emph{overlapping}.
        On synthetic data for additive noise levels increasing from $0\%$ to $50\%$, we show mean as line and standard deviation as shade of \emph{similarity}, \emph{recall} w.r.t. the target matrix, and \emph{\recallgt}\ w.r.t. the noise-free ground-truth tiles, for \binaps, \asso, \grecond, \LOM, \sofa, \nmf, \pimp, and \ourmethod.
    }
    \label{fig:noise-overlapping}
\end{figure*}

In this section, we provide additional results on our synthetic experiments in Sec.~\ref{sec:noise} and Fig.~\ref{fig:noise}.

In our synthetic experiments in Sec.~\ref{sec:noise}, we simulate data by generating \emph{non-overlapping} tiles using rejection sampling.
That is, before we place the next randomly drawn tile into our matrix, we check for overlap with past placements. 
If we detect an overlap, we reject, redraw, and repeat, until we placed the desired number of tiles into our matrix.
In the following, we allow \emph{overlapping} tiles, for which we simply omit the rejection step laid out above. 
To depict results on harder-to-separate data, we generate synthetic matrices as described in Sec.~\ref{sec:experiments}, however, this time, allowing tiles to overlap arbitrarily.
In Fig.~\ref{fig:noise-overlapping}, we show similarity, recall, and \recallgt for the \emph{overlapping case},
observing a similar behavior to Fig.~\ref{fig:noise} across the board.
Again noticeable is the surprisingly good performance of rounded \nmf reconstructions, outperforming \asso, \grecond, \sofa, and \binaps by a large margin. 
Overall, \pimp and \ourmethod outperform \asso, \grecond, \LOM, \sofa, \binaps, and \nmf across varying noise levels in similarity, recall, and \recallgt.
Fig.~\ref{fig:noise} and Fig.~\ref{fig:noise-overlapping} show that \ourmethod, which does not use any post-processing, achieves best-in-class results for \emph{overlapping} and \emph{non-overlapping} tiles, on par with the strongest competitor, which relies heavily on post-processing.

\section{Additional Results on the Performance of \ourmethod on Synthetic Data with Varying Densities}\label{apx:densities}

\begin{figure*}[t] 
    \centering
    \includegraphics{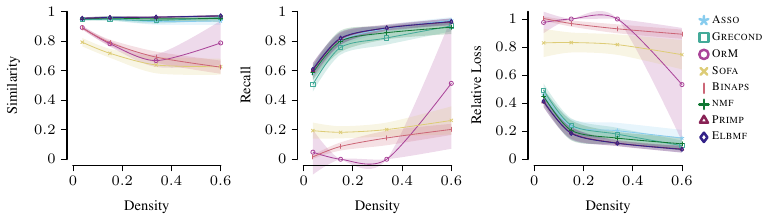}
    \caption{
        \ourmethod reconstructs the low-noise synthetic high- and low-density matrices well and consistently so. 
        On synthetic data with fixed additive noise level of as low as $5\%$, and increasing density, we show mean as line and standard deviation as shade of \emph{similarity}, \emph{recall}, and \emph{relative loss} w.r.t. the target matrix, for \binaps, \asso, \grecond, \LOM, \sofa, \nmf, \pimp, and \ourmethod.
    }
    \label{fig:density-low-noise}
\end{figure*}

Continuing our experiments from Sec.~\ref{sec:densities} and Fig.~\ref{fig:density-high-noise},
we ask whether our observations carry over to a low-noise scenario, in which \asso and \grecond performance improves significantly (see Fig.~\ref{fig:noise}).

For this, we study the effects of varying densities under a low noise level of only $5\%$. 
As tiny tiles are hard to distinguish from noise, we see an overall improvement with increasing density, regardless of the method.
With less noise, \asso, \grecond, and \nmf improve significantly in comparison to their performance under more noise (Fig.~\ref{fig:density-high-noise}).
They, however, are still outperformed by \pimp and \ourmethod in recall and loss. 
The similarities of \asso, \grecond, \nmf, \pimp, and \ourmethod are close to $1$, whereas \sofa, \binaps, and \LOM exhibit lower similarity with increasing density.
From Fig.~\ref{fig:density-low-noise} and Fig.~\ref{fig:density-high-noise}, we see that \ourmethod performs consistently well across varying densities, regardless of the noise level.

\end{document}